\documentclass[11pt]{article}
\usepackage{amsmath}
\usepackage{acl}

\usepackage{times}
\usepackage{latexsym}
\usepackage[T1]{fontenc}
\usepackage[utf8]{inputenc}
\usepackage{microtype}
\usepackage{inconsolata}
\usepackage{graphicx}
\usepackage{booktabs}
\usepackage{multirow}
\usepackage{tikz}
\usetikzlibrary{shapes.geometric, arrows.meta, positioning, fit, backgrounds}
\newcommand\blfootnote[1]{%
  \begingroup
  \renewcommand\thefootnote{}\footnote{#1}%
  \addtocounter{footnote}{-1}%
  \endgroup
}

\title{CVSS-X: A Multilingual Speech-to-Speech Translation Corpus for 28 Languages}

\author{
  Lucas Rafael Stefanel Gris$^{1}$,
  Alef Iury Siqueira Ferreira$^{1}$,
  Frederico Santos de Oliveira$^{2}$, \\
  \textbf{Augusto Seben da Rosa}$^{3}$,
  \textbf{Alexandre Costa Ferro Filho}$^{1}$,
  \textbf{Arlindo Rodrigues Galv{\~a}o Filho}$^{1}$, \\
  \textbf{Anderson da Silva Soares}$^{1}$ \\
  $^{1}$Federal University of Goiás, Goiânia, Brazil \\
  $^{2}$Federal University of Mato Grosso, Cuiabá, Brazil \\
  $^{3}$São Paulo State University, São Paulo, Brazil \\
  \texttt{lucas.gris@discente.ufg.br}
}

\begin{document}

\maketitle

\begin{abstract}
We introduce CVSS-X, a large-scale synthetic speech-to-speech translation corpus that extends CVSS by reversing the translation direction.
While CVSS translates from 21 languages into English, CVSS-X enables translation from English into 28 target languages spanning 12 language families.
The corpus comprises approximately 240,000 parallel speech pairs per language, totaling over 16,000 hours, eight times larger than CVSS.
We provide two variants: CVSS-X-C with two canonical voices per language, and CVSS-X-T with cross-lingual voice cloning, both fully generated.
Evaluation shows comparable translation quality to CVSS with consistent performance across typologically diverse languages.
Combined with CVSS, this enables research on bidirectional and multilingual speech-to-speech translation.
The code is available at \url{https://github.com/ErmisAI/XVSS-X} and the dataset under CC-BY-NC 4.0 license at \url{https://huggingface.co/datasets/lgris/XVSS-X}.
\end{abstract}

\blfootnote{Accepted at the SALMA Workshop (2nd Edition) @ EMNLP 2026 (Non-archival). OpenReview: \url{https://openreview.net/forum?id=HQw3teisL9}}

\section{Introduction}

Speech-to-speech translation (S2ST) is a key technology for breaking down communication barriers between speakers of different languages.
Traditionally, S2ST systems rely on a cascade of automatic speech recognition (ASR), text-based machine translation (MT), and text-to-speech (TTS) synthesis \citep{Waibel1991}.
Recently, direct S2ST approaches that bypass intermediate text representations have gained significant attention, including end-to-end models such as Translatotron \citep{Jia2019b,Jia2022translatotron2} and methods based on discrete speech units \citep{Lee2022,Popuri2022}.

The release of the CVSS corpus \citep{Jia2022cvss} marked a turning point, providing the first large-scale, publicly available parallel speech corpus and enabling substantial progress in direct S2ST research.
However, CVSS is restricted to a many-to-one setting, translating from 21 source languages exclusively into English.
This limitation constrains the development of more inclusive multilingual S2ST technologies, as researchers working on translation \emph{from} English or between non-English language pairs still face a scarcity of suitable training data.

In this paper, we introduce \textbf{CVSS-X}, a massively multilingual speech-to-speech translation corpus that complements and extends the original CVSS by reversing the translation direction.
While CVSS translates \emph{into} English from 21 source languages, CVSS-X translates \emph{from} English into 28 target languages spanning 12 language families, including Germanic, Romance, Slavic, CJK, and others.
CVSS-X is derived from English Common Voice recordings \citep{Ardila2020}, with translations generated using NLLB-200 \citep{Costa-jussa2022} and synthesized via OmniVoice \citep{zhu2026omnivoice}, a state-of-the-art (SOTA) multilingual TTS system with voice cloning capabilities.

Following the design of CVSS, we provide two variants: \textbf{CVSS-X-C} (Canonical), where target speech is synthesized using two fixed reference voices per language, and \textbf{CVSS-X-T} (Transferred), where the voice characteristics of the original English speaker are preserved through cross-lingual voice cloning.
The combination of CVSS and CVSS-X enables, for the first time, large-scale research on bidirectional and multilingual S2ST, supporting translation not only to and from English, but also between arbitrary language pairs using English as a pivot.

\section{Related Work}

Table~\ref{tab:corpora} summarizes publicly available corpora for S2ST research.
True parallel speech corpora remain rare due to the high cost of collecting aligned utterances across languages.
CVSS \citep{Jia2022cvss} was the first large-scale public S2ST corpus, but is limited to many-to-one translation (X$\rightarrow$EN) with 21 source languages.
SpeechMatrix \citep{Duquenne2023} provides massive scale (418K hours) through automatic speech mining from European Parliament recordings, though it covers only 17 European languages and alignment quality is inherently approximate.
The SeamlessM4T project \citep{Communication2023seamless} introduced SeamlessAlign, expanding coverage to 37 languages, but only metadata is publicly released, requiring users to reconstruct the dataset from Common Crawl archives.

\begin{table}[t]
\centering
\caption{Comparison of S2ST corpora. Dirs: number of supported translation directions. Type: \textit{Real} (human recordings), \textit{Mined} (automatically aligned), \textit{Synth} (TTS-generated).}
\label{tab:corpora}
\resizebox{\columnwidth}{!}{
\begin{tabular}{lccccc}
\toprule
\textbf{Dataset} & \textbf{Dirs} & \textbf{Langs} & \textbf{Hours} & \textbf{Type} & \textbf{License} \\
\midrule
Fisher ES-EN & 2 & 2 & 127 & Real & LDC \\
VoxPopuli & 210 & 15 & 17.3K & Real & CC0 \\
SpeechMatrix & 136 & 17 & 418K & Mined & CC-BY-NC \\
SeamlessAlign & -- & 37 & 29K & Mined & CC-BY-NC$^\dagger$ \\
\midrule
CVSS-C/T & 21 & 21 & 1.9K & Synth & CC-BY \\
\textbf{CVSS-X-C/T} & \textbf{28} & \textbf{28} & \textbf{16K} & \textbf{Synth} & \textbf{CC-BY-NC} \\
\bottomrule
\multicolumn{6}{l}{\footnotesize $^\dagger$Metadata only; requires reconstruction from Common Crawl.} \\
\end{tabular}%
}
\end{table}

CVSS-X addresses these limitations by providing a fully downloadable synthetic corpus that enables one-to-many translation (EN$\rightarrow$X) across 28 languages from 12 typological families (grouped into 7 macro-categories).
Unlike mined corpora, CVSS-X offers perfect sentence-level alignment since source and target speech are generated from verified parallel text.

\section{CVSS-X Dataset}

\subsection{Data Sources}

  \textbf{Source Data.}
  We construct CVSS-X using English source utterances from Common Voice version 17 \citep{Ardila2020}.
  To establish direct symmetry with CVSS \citep{Jia2022cvss}, which synthesized English target speech corresponding to CoVoST 2 English transcripts, we align Common Voice version 17 recordings with the original Common Voice version 4 utterances used by CVSS/CoVoST 2 via normalized text matching.
  This recovered 240,192 of the original 264,037 English utterances (91.0\%) with verified human source audio.
  Since the matched dev split contained only 871 samples, we supplemented it by randomly sampling from the training split to reach exactly 10,000 samples.
  The final splits are: \textbf{train} (222,349), \textbf{dev} (10,000), and \textbf{test} (7,843) samples per target language.

  \textbf{Text Translation.}

  Source transcripts are translated using the distilled NLLB-200 model\footnote{\url{https://huggingface.co/facebook/nllb-200-distilled-600M}}~\citep{Costa-jussa2022}, selected after benchmarking seven candidate translation models on a closed evaluation set. Translation quality was assessed through an LLM-as-a-judge protocol over 100 EN$\rightarrow$PT samples, with each output rated on a 1--10 scale for accuracy, fluency, and terminology preservation.

  NLLB-200 was ultimately selected due to its strong performance in this evaluation, together with its broad coverage of languages and language families, making it particularly suitable for the multilingual setting considered in this work.

\textbf{Speech Synthesis.}
Target speech is synthesized using OmniVoice \citep{zhu2026omnivoice}, a SOTA multilingual TTS system with zero-shot voice cloning capabilities.
OmniVoice was selected for its ability to perform cross-lingual voice cloning, synthesizing speech in a target language from a voice reference spoken in a different source language. This capability is particularly important for CVSS-X-T, where the available voice reference is English speech, as it helps preserve speaker identity while minimizing the transfer of English-specific accent characteristics to the synthesized target-language speech.

\subsection{Generation Pipeline}

Figure~\ref{fig:pipeline} illustrates the CVSS-X generation pipeline.
We generate two corpus variants:

\begin{figure}[t]
\centering
\begin{tikzpicture}[
    node distance=0.5cm and 0.3cm,
    box/.style={rectangle, draw, rounded corners, minimum height=0.85cm, minimum width=1.5cm, align=center, font=\scriptsize},
    data/.style={box, fill=blue!15},
    process/.style={box, fill=orange!20},
    output/.style={box, fill=green!15},
    arrow/.style={-{Stealth[length=2mm]}, thick},
    label/.style={font=\tiny, align=center}
]
\node[data] (cv) {Common Voice\\English (v17)};
\node[process, right=of cv] (nllb) {NLLB-200\\Translation};
\node[process, right=of nllb] (tts) {OmniVoice\\TTS};
\node[output, right=of tts] (cvss) {CVSS-X\\Dataset};

\draw[arrow] (cv) -- (nllb);
\draw[arrow] (nllb) -- (tts);
\draw[arrow] (tts) -- (cvss);

\node[label, below=0.15cm of cv] {240K utterances\\1,153 hours};
\node[label, below=0.15cm of nllb] {EN $\rightarrow$ 28 langs};
\node[label, below=0.15cm of tts] {Canonical +\\Voice Cloning};
\node[label, below=0.15cm of cvss] {16K hours\\28 languages};
\end{tikzpicture}
\caption{CVSS-X dataset creation pipeline.}
\label{fig:pipeline}
\end{figure}
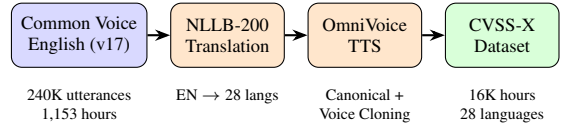

\textbf{CVSS-X-C (Canonical):} Two fixed reference voices per language (one male, one female), created using ElevenLabs'\footnote{https://elevenlabs.io} voice design feature, with synthetic speakers with neutral accents optimized for clarity.
Voice selection is based on source speaker gender metadata: samples with \texttt{gender=male}, \texttt{empty}, or \texttt{other} use the male voice; \texttt{gender=female} samples use the female voice.
This results in 81.4\% male and 18.6\% female distribution, doubling voice diversity compared to CVSS-C's single canonical voice.

\textbf{CVSS-X-T (Timbre-transferred):} Zero-shot voice cloning conditioned on the source English audio, preserving speaker characteristics across languages.
Approximately 3.4\% of samples are skipped due to insufficient signal for voice cloning.

\subsection{Target Languages}

CVSS-X covers 28 languages spanning 12 distinct phylogenetic and typological language families (Romance, Germanic, Slavic, Sinitic, Japonic, Koreanic, Uralic, Indo-Iranian, Hellenic, Semitic, Turkic, and Austroasiatic/Austronesian/Kra-Dai), grouped into 7 macro-categories in Table~\ref{tab:languages} to facilitate comparative analysis.

\begin{table}[t]
\centering
\small
\setlength{\tabcolsep}{4pt}
\caption{CVSS-X target languages by family.}
\label{tab:languages}
\begin{tabular}{lcl}
\toprule
\textbf{Family} & \textbf{N} & \textbf{Languages} \\
\midrule
Romance & 6 & PT, ES, FR, IT, RO, CA \\
Germanic & 5 & DE, NL, SV, DA, NO \\
Slavic & 4 & RU, PL, CS, UK \\
CJK & 3 & ZH, JA, KO \\
Uralic & 2 & FI, HU \\
Indo-Iranian & 2 & HI, FA \\
Other & 6 & EL, HE, TR, TH, ID, VI \\
\bottomrule
\end{tabular}
\end{table}

\subsection{Corpus Statistics}

Table~\ref{tab:statistics} summarizes corpus statistics.
Both CVSS-X-C and CVSS-X-T are fully generated, totaling over 16,000 hours of parallel speech.

\begin{table}[t]
\centering
\small
\caption{CVSS-X corpus statistics.}
\label{tab:statistics}
\begin{tabular}{lr}
\toprule
\textbf{Metric} & \textbf{Value} \\
\midrule
Source language & English \\
Target languages & 28 \\
Language families & 7 \\
Samples per language & 240,192 \\
Total translated pairs & 6,725,376 \\
\midrule
\textbf{CVSS-X-C} & \\
\quad Duration & $\sim$6,730 hours \\
\quad Avg. utterance & 3.6 seconds \\
\textbf{CVSS-X-T} &  \\
\quad Duration & $\sim$9,340 hours \\
\quad Avg. utterance & 5.0 seconds \\
\midrule
\textbf{Total duration} & $\sim$16,070 hours \\
\bottomrule
\end{tabular}
\end{table}

\section{Dataset Quality Evaluation}

We evaluate CVSS-X quality and compare with the original CVSS corpus.
Note that direct comparison has limitations since CVSS evaluates X$\rightarrow$EN while CVSS-X evaluates EN$\rightarrow$X.

\subsection{Evaluation Setup}

We evaluate a stratified random sample of 200 utterances per language from the dev set (5,600 samples per variant).
This sample size was determined via power analysis: given observed standard deviations of $\sigma \approx 0.5$ for UTMOS, 200 samples yields 95\% confidence intervals of $\pm 0.07$, sufficient to detect meaningful differences.
To evaluate the acoustic intelligibility and text preservation of the synthesized target speech, we perform Whisper large-v3 ASR \citep{Radford2023whisper} and compute Word/Character Error Rate (WER/CER) and ASR-BLEU against the translated text prompt (measuring TTS-to-ASR round-trip fidelity).
For unsegmented languages (ZH, JA, TH), standard whitespace-based SacreBLEU produces zero; we therefore employ language-specific tokenizers (\textit{jieba}, \textit{fugashi}, and \textit{pythainlp}) and report Character Error Rate (CER) and chrF2++~\cite{popovic-2015-chrf, popovic-2016-chrf, popovic-2017-chrf}.
Naturalness is measured with UTMOS \citep{Saeki2022utmos}, and speaker similarity with ECAPA-TDNN \citep{Desplanques2020ecapa}.
To enable fair comparison with CVSS, we re-evaluate the original corpus using the same models and sample size.

\subsection{Results}

Table~\ref{tab:results_family} shows results by language family, and Table~\ref{tab:results_summary} provides an overall comparison.

\begin{table}[t]
\centering
\scriptsize 
\setlength{\tabcolsep}{2pt} 
\caption{Evaluation by language family (CVSS-X).}
\label{tab:results_family}
\begin{tabular*}{\columnwidth}{l @{\extracolsep{\fill}} cccccc}
\toprule
& \multicolumn{2}{c}{\textbf{WER/CER}} & \multicolumn{2}{c}{\textbf{ASR-BLEU}} & \multicolumn{2}{c}{\textbf{UTMOS}} \\
\cmidrule(lr){2-3} \cmidrule(lr){4-5} \cmidrule(lr){6-7}
\textbf{Family} & \textbf{C} & \textbf{T} & \textbf{C} & \textbf{T} & \textbf{C} & \textbf{T} \\
\midrule
Romance (6) & 5.8 & 8.2 & 90.3 & 88.0 & 3.54 & 3.27 \\
Germanic (5) & 9.5 & 12.2 & 85.3 & 81.1 & 3.62 & 3.27 \\
Slavic (4) & 7.0 & 7.7 & 86.9 & 85.9 & 3.48 & 3.17 \\
CJK (3)$^*$ & 5.0 & 10.2 & 75.2 & 67.9 & 3.56 & 3.20 \\
Uralic (2) & 10.3 & 14.3 & 84.0 & 78.3 & 3.54 & 3.21 \\
Indo-Iranian (2) & 22.5 & 23.0 & 63.7 & 60.0 & 3.61 & 3.20 \\
Other (6) & 24.8 & 24.8 & 65.0 & 64.9 & 3.53 & 3.14 \\
\midrule
\textbf{Average} & \textbf{12.1} & \textbf{14.1} & \textbf{82.4} & \textbf{79.4} & \textbf{3.55} & \textbf{3.21} \\
\bottomrule
\multicolumn{7}{l}{\footnotesize $^*$CER reported; BLEU w/ specific tokenizers.} \\
\end{tabular*}
\end{table}

\begin{table}[t]
\centering
\small
\caption{Overall comparison with CVSS (re-evaluated with same pipeline).}
\label{tab:results_summary}
\begin{tabular}{lcccc}
\toprule
\textbf{Metric} & \textbf{X-C} & \textbf{X-T} & \textbf{CVSS-C} & \textbf{CVSS-T} \\
\midrule
UTMOS (1--5) & 3.55 & 3.21 & 4.43 & 3.61 \\
ASR-BLEU & 82.4 & 79.4 & 94.2 & 93.8 \\
WER/CER (\%) & 12.1 & 14.1 & 3.5 & 4.0 \\
Spk. Similarity & -- & 0.607 & -- & -- \\
\bottomrule
\end{tabular}
\end{table}

\textbf{TTS quality gap.}
CVSS-X-C achieves lower UTMOS (3.55) compared to CVSS-C (4.43).
This gap stems from task complexity: CVSS synthesizes only English using PnG NAT trained on high-quality LibriTTS data, while CVSS-X must synthesize 28 typologically diverse languages with a single multilingual model.
The consistent UTMOS across language families (3.48--3.62) suggests OmniVoice provides uniform quality rather than excelling in some languages at the expense of others.

\textbf{Voice cloning quality.}
CVSS-X-T achieves UTMOS of 3.21, a modest reduction from CVSS-X-C (3.55).
This gap is smaller than expected given that OmniVoice preserves acoustic characteristics from the source audio, including potential artifacts from crowdsourced recordings.
Speaker similarity of 0.607 indicates good voice preservation, with Germanic languages achieving higher similarity (0.648) than Slavic (0.567), possibly due to phonetic proximity to English.

\textbf{TTS intelligibility and fidelity.}
Romance and Slavic families achieve the highest BLEU scores (88--90) and lowest error rates (WER $<8\%$), demonstrating high synthesis fidelity across OmniVoice.
Indo-Iranian and Other families show lower scores (60--65) driven by non-Latin scripts such as Hebrew (ASR-BLEU=46.1) and complex tonal dynamics in Thai.
When using language-specific word tokenizers, unsegmented languages achieve strong fidelity: Thai reaches word-tokenized BLEU of 49.1 (C) and 47.8 (T), while Chinese (ZH) and Japanese (JA) achieve tokenized BLEU of 87.6/70.4 and 88.9/85.7, respectively.
The higher WER in the ``Other'' family (24.8\%) is primarily driven by Thai (77.0\% WER without tone diacritics in Whisper) and Hebrew (37.4\% WER).

\section{Conclusion}

We introduced CVSS-X, a large-scale synthetic S2ST corpus that reverses the translation direction of CVSS, enabling EN$\rightarrow$X translation into 28 target languages across 12 language families.
Both variants are fully generated: CVSS-X-C (6,730 hours) and CVSS-X-T (9,340 hours), totaling over 16,000 hours of parallel speech---eight times larger than CVSS.
Combined with CVSS, this enables bidirectional translation and, through English as pivot, arbitrary X$\rightarrow$Y language pairs.

Future versions will address current limitations: regenerating translations with TranslateGemma-12B would improve fidelity and enable Apache 2.0 licensing; incorporating Common Voice v26
  (significantly more speakers) and Common Voice Spontaneous Speech 4.0 (natural conversational recordings) would increase diversity and scale while providing more realistic voice cloning scenarios.
  We also plan to train and evaluate baseline S2ST models to validate the dataset.

\section*{Limitations}

The current release has several limitations: (1) translation quality depends on NLLB-200, which, despite verified fidelity, may introduce errors for low-resource languages; (2) while ASR-BLEU measures TTS acoustic intelligibility against prompt text, cross-validation against human-authored references (such as CoVoST 2 English-to-15 subsets) remains an ongoing benchmark; (3) the CC-BY-NC license (inherited from NLLB-200) restricts commercial applications; (4) comparison between CVSS (X$\rightarrow$\text{EN}) and CVSS-X (\text{EN}$\rightarrow$ X) is inherently asymmetric; and (5) automated MOS predictors (UTMOS) may exhibit variance across non-Western phonologies, necessitating human perceptual validation in future iterations.

\section*{Acknowledgments} 

This work has been funded by the project Research and Development of Genese Digital: Scaling Interactive and Culturally Adapted Digital Humans with Generative AI, supported by the Advanced Knowledge Center in Immersive Technologies (AKCIT), with financial resources from the PPI IoT/Manufatura 4.0 / PPI HardwareBR of the MCTI, grant number 057/2023, signed with EMBRAPII. The authors also acknowledge the support and contributions of Huglabs and Ermis.ai.

\bibliography{refs}

@inproceedings{Waibel1991,
    author    = {Alex Waibel and Ajay N. Jain and Arthur E. McNair and Hiroaki Saito and Alexander G. Hauptmann and Joe Tebelskis},
    title     = {{JANUS}: A Speech-to-Speech Translation System Using Connectionist and Symbolic Processing Strategies},
    booktitle = {Proceedings of the IEEE International Conference on Acoustics, Speech and Signal Processing (ICASSP)},
    year      = {1991},
    pages     = {793--796},
  }

@inproceedings{Jia2019b,
    author    = {Ye Jia and Ron J. Weiss and Fadi Biadsy and Wolfgang Macherey and Melvin Johnson and Zhifeng Chen and Yonghui Wu},
    title     = {Direct Speech-to-Speech Translation with a Sequence-to-Sequence Model},
    booktitle = {Proceedings of Interspeech},
    year      = {2019},
  }

@inproceedings{Jia2022translatotron2,
    author    = {Ye Jia and Michelle Tadmor Ramanovich and Tal Remez and Roi Pomerantz},
    title     = {Translatotron 2: High-Quality Direct Speech-to-Speech Translation with Voice Preservation},
    booktitle = {Proceedings of the International Conference on Machine Learning (ICML)},
    year      = {2022},
  }

@inproceedings{Lee2022,
    author    = {Ann Lee and Peng-Jen Chen and Changhan Wang and Jiatao Gu and Xutai Ma and Adam Polyak and Yossi Adi and Qing He and Yun Tang and Juan Pino and Wei-Ning Hsu},
    title     = {Direct Speech-to-Speech Translation with Discrete Units},
    booktitle = {Proceedings of the Annual Meeting of the Association for Computational Linguistics (ACL)},
    year      = {2022},
  }

@inproceedings{Popuri2022,
    author    = {Sravya Popuri and Peng-Jen Chen and Changhan Wang and Juan Pino and Yossi Adi and Jiatao Gu and Wei-Ning Hsu and Ann Lee},
    title     = {Enhanced Direct Speech-to-Speech Translation Using Self-supervised Pre-training and Data Augmentation},
    booktitle = {Proceedings of Interspeech},
    year      = {2022},
  }

@inproceedings{Jia2022cvss,
    author    = {Ye Jia and Michelle Tadmor Ramanovich and Quan Wang and Heiga Zen},
    title     = {{CVSS} Corpus and Massively Multilingual Speech-to-Speech Translation},
    booktitle = {Proceedings of the Language Resources and Evaluation Conference (LREC)},
    year      = {2022},
    note      = {arXiv:2201.03713},
  }

@inproceedings{Duquenne2023,
    author    = {Paul-Ambroise Duquenne and Hongyu Gong and Ning Dong and Jingfei Du and Ann Lee and Vedanuj Goswami and Changhan Wang and Juan Pino and Beno{\^i}t Sagot and Holger Schwenk},
    title     = {{SpeechMatrix}: A Large-Scale Mined Corpus of Multilingual Speech-to-Speech Translations},
    booktitle = {Proceedings of the Annual Meeting of the Association for Computational Linguistics (ACL)},
    year      = {2023},
    pages     = {16251--16269},
  }

@article{Communication2023seamless,
    author    = {{Seamless Communication} and Lo{\"i}c Barrault and Yu-An Chung and Mariano Coria Meglioli and David Dale and Ning Dong and Paul-Ambroise Duquenne and Hady ElSahar and Hongyu Gong and Kevin Heffernan and John Hoffman and
  Christopher Klaiber and Pengwei Li and Daniel Licht and Jean Maillard and Alice Rakotoarison and Kaushik Ram Sadagopan and Guillaume Wenzek and Ethan Ye and Bapi Akula and Peng-Jen Chen and Naji El Hachem and Brian Ellis and Gabriel
  Mejia Gonzalez and Justin Haaheim and Prangthip Hansanti and Russ Howes and Bernie Huang and Min-Jae Hwang and Hirofumi Inaguma and Somya Jain and Elahe Kalbassi and Amanda Kallet and Ilia Kulikov and Janice Lam and Daniel Li and
  Xutai Ma and Ruslan Mavlyutov and Benjamin Peloquin and Mohamed Ramadan and Abinesh Ramakrishnan and Anna Sun and Kevin Tran and Tuan Tran and Igor Tufanov and Vish Vogeti and Carleigh Wood and Yilin Yang and Bokai Yu and Pierre
  Andrews and Can Balioglu and Marta R. Costa-juss{\`a} and Onur {\c{C}}elebi and Maha Elbayad and Cynthia Gao and Francisco Guzm{\'a}n and Justine Kao and Ann Lee and Alexandre Mourachko and Juan Pino and Sravya Popuri and Christophe
  Ropers and Safiyyah Saleem and Holger Schwenk and Paden Tomasello and Changhan Wang and Jeff Wang and Skyler Wang},
    title     = {{SeamlessM4T}: Massively Multilingual \& Multimodal Machine Translation},
    journal   = {arXiv preprint arXiv:2308.11596},
    year      = {2023},
  }

@inproceedings{Ardila2020,
    author    = {Rosana Ardila and Megan Branson and Kelly Davis and Michael Henretty and Michael Kohler and Josh Meyer and Reuben Morais and Lindsay Saunders and Francis M. Tyers and Gregor Weber},
    title     = {Common Voice: A Massively-Multilingual Speech Corpus},
    booktitle = {Proceedings of the Language Resources and Evaluation Conference (LREC)},
    year      = {2020},
  }

@article{Costa-jussa2022,
    author    = {Marta R. Costa-juss{\`a} and James Cross and Onur {\c{C}}elebi and Maha Elbayad and Kenneth Heafield and Kevin Heffernan and Elahe Kalbassi and Janice Lam and Daniel Licht and Jean Maillard and Anna Sun and Skyler Wang
  and Guillaume Wenzek and Al Youngblood and Bapi Akula and Loic Barrault and Gabriel Mejia Gonzalez and Prangthip Hansanti and John Hoffman and Semarley Jarrett and Kaushik Ram Sadagopan and Dirk Rowe and Shannon Spruit and Chau Tran
  and Pierre Andrews and Necip Fazil Ayan and Shruti Bhosale and Sergey Edunov and Angela Fan and Cynthia Gao and Vedanuj Goswami and Francisco Guzm{\'a}n and Philipp Koehn and Alexandre Mourachko and Christophe Ropers and Safiyyah
  Saleem and Holger Schwenk and Jeff Wang},
    title     = {No Language Left Behind: Scaling Human-Centered Machine Translation},
    journal   = {arXiv preprint arXiv:2207.04672},
    year      = {2022},
  }

@article{zhu2026omnivoice,
      title={OmniVoice: Towards Omnilingual Zero-Shot Text-to-Speech with Diffusion Language Models},
      author={Zhu, Han and Ye, Lingxuan and Kang, Wei and Yao, Zengwei and Guo, Liyong and Kuang, Fangjun and Han, Zhifeng and Zhuang, Weiji and Lin, Long and Povey, Daniel},
      journal={arXiv preprint arXiv:2604.00688},
      year={2026}
}

@inproceedings{Desplanques2020ecapa,
    author    = {Brecht Desplanques and Jenthe Thienpondt and Kris Demuynck},
    title     = {{ECAPA-TDNN}: Emphasized Channel Attention, Propagation and Aggregation in {TDNN} Based Speaker Verification},
    booktitle = {Proceedings of Interspeech},
    year      = {2020},
    pages     = {3830--3834},
  }

@article{Radford2023whisper,
    author    = {Alec Radford and Jong Wook Kim and Tao Xu and Greg Brockman and Christine McLeavey and Ilya Sutskever},
    title     = {Robust Speech Recognition via Large-Scale Weak Supervision},
    journal   = {Proceedings of the International Conference on Machine Learning (ICML)},
    year      = {2023},
  }

@article{Saeki2022utmos,
    author    = {Takaaki Saeki and Detai Xin and Wataru Nakata and Tomoki Koriyama and Shinnosuke Takamichi and Hiroshi Saruwatari},
    title     = {{UTMOS}: {UTokyo-SaruLab} System for {VoiceMOS} Challenge 2022},
    journal   = {arXiv preprint arXiv:2204.02152},
    year      = {2022},
  }

@inproceedings{popovic-2015-chrf,
    title = "chr{F}: character n-gram {F}-score for automatic {MT} evaluation",
    author = "Popovi{\'c}, Maja",
    editor = "Bojar, Ond{\v{r}}ej  and
      Chatterjee, Rajan  and
      Federmann, Christian  and
      Haddow, Barry  and
      Hokamp, Chris  and
      Huck, Matthias  and
      Logacheva, Varvara  and
      Pecina, Pavel",
    booktitle = "Proceedings of the Tenth Workshop on Statistical Machine Translation",
    month = sep,
    year = "2015",
    address = "Lisbon, Portugal",
    publisher = "Association for Computational Linguistics",
    url = "https://aclanthology.org/W15-3049/",
    doi = "10.18653/v1/W15-3049",
    pages = "392--395"
}

@inproceedings{popovic-2016-chrf,
    title = "chr{F} deconstructed: beta parameters and n-gram weights",
    author = "Popovi{\'c}, Maja",
    editor = {Bojar, Ond{\v{r}}ej  and
      Buck, Christian  and
      Chatterjee, Rajen  and
      Federmann, Christian  and
      Guillou, Liane  and
      Haddow, Barry  and
      Huck, Matthias  and
      Yepes, Antonio Jimeno  and
      N{\'e}v{\'e}ol, Aur{\'e}lie  and
      Neves, Mariana  and
      Pecina, Pavel  and
      Popel, Martin  and
      Koehn, Philipp  and
      Monz, Christof  and
      Negri, Matteo  and
      Post, Matt  and
      Specia, Lucia  and
      Verspoor, Karin  and
      Tiedemann, J{\"o}rg  and
      Turchi, Marco},
    booktitle = "Proceedings of the First Conference on Machine Translation: Volume 2, Shared Task Papers",
    month = aug,
    year = "2016",
    address = "Berlin, Germany",
    publisher = "Association for Computational Linguistics",
    url = "https://aclanthology.org/W16-2341/",
    doi = "10.18653/v1/W16-2341",
    pages = "499--504"
}

@inproceedings{popovic-2017-chrf,
    title = "chr{F}++: words helping character n-grams",
    author = "Popovi{\'c}, Maja",
    editor = "Bojar, Ond{\v{r}}ej  and
      Buck, Christian  and
      Chatterjee, Rajen  and
      Federmann, Christian  and
      Graham, Yvette  and
      Haddow, Barry  and
      Huck, Matthias  and
      Yepes, Antonio Jimeno  and
      Koehn, Philipp  and
      Kreutzer, Julia",
    booktitle = "Proceedings of the Second Conference on Machine Translation",
    month = sep,
    year = "2017",
    address = "Copenhagen, Denmark",
    publisher = "Association for Computational Linguistics",
    url = "https://aclanthology.org/W17-4770/",
    doi = "10.18653/v1/W17-4770",
    pages = "612--618"
}

\end{document}